\documentclass[a4paper]{article}

\usepackage{INTERSPEECH_v2}
\usepackage{makecell}

\title{A practical approach to dialogue response generation in closed domains}
\name{Yichao Lu$^1$, Phillip Keung$^1$, Shaonan Zhang $^1$, Jason Sun$^1$, Vikas Bhardwaj$^1$}
\address{
  $^1$Amazon Inc., USA}
\email{yichaolu@amazon.com, keung@amazon.com}

\begin{document}

\maketitle
\begin{abstract}

We describe a prototype dialogue response generation model for the customer service domain at Amazon. The model, which is trained in a weakly supervised fashion, measures the similarity between customer questions and agent answers using a dual encoder network, a Siamese-like neural network architecture. Answer templates are extracted from embeddings derived from past agent answers, without turn-by-turn annotations. Responses to customer inquiries are generated by selecting the best template from the final set of templates. We show that, in a closed domain like customer service, the selected templates cover $>$70\% of past customer inquiries. Furthermore, the relevance of the model-selected templates is significantly higher than templates selected by a standard tf-idf baseline.

\end{abstract}

\section{Introduction}

Millions of shoppers contact Amazon's customer service department every year, where customers may choose between telephone, online chat, or email channels. Most customers will contact Amazon via telephone, which is an especially labor-intensive form of communication. The need for agent labor is highly seasonal, and hiring more agents requires significant ramp-up time for training. Furthermore, Amazon's order volume increases significantly year-over-year, which makes scaling customer service sub-linearly with order volume especially crucial.

Machine learning and dialogue generation provide an opportunity to make existing agents more efficient, and may allow for the total automation of issue resolution (at least for a select subset of issues.) To that end, we present the first steps towards a practical dialogue system for the customer service domain at Amazon. In this work, we focus solely on the response generation module of such a system: given a customer inquiry, generate the text of the response that most likely answers the question asked. An effective dialogue system would automate the handling of a large percentage of customer interactions, potentially generating significant savings in labor costs, reducing the perceived response times for customers, and allowing customer service to scale better with increasing demand.

For well-established domains like customer service, large human-generated corpuses already exist. Indeed, Amazon's internal online chat corpus is a rich source of data for building a response generation model, independently of transcribed speech from past phone calls. Amazon's chat corpus also contains customer-selected issue labels, order-related entities, etc. We will use this corpus for model training and experiments.

While open domain dialogue generation remains a topic of ongoing research, we hypothesize that in closed domains like our own, a finite set of response templates would cover the vast majority of interactions.  In fact, customer service agents across Amazon already use a collection of `blurbs' which they copy and paste as replies. However, these blurbs are not centrally managed, can be particular to each agent, are unannotated, and number in the thousands. In practice, due to the overhead of searching for the right blurb, each agent only uses a handful of blurbs regularly.

Therefore, we approach response generation as a two-fold problem: determining the templates that should be created based on past agent replies to customer questions, and choosing the correct template as the response to an inquiry.

A template-based approach addresses some issues that affect existing dialogue generation systems, namely relevance, text quality, diversity, and (for goal-oriented systems) the need for annotated customer intents. The templates that we extract can be filtered for high relevance and specificity, corrected for consistency of tone, and enriched with the addition of slots for customer profile metadata and other forms of context. Furthermore, the use of fixed templates allows us to better tune text-to-speech systems to produce more natural-sounding speech.

We built a prototype response generation model based on online chats between customers and agents, and evaluated it against an random sample of past chat conversations. We showed that the selected templates cover a large portion of past customer inquiries, and that human evaluators (who are customer service agents) preferred the model-selected templates to the templates retrieved by a tf-idf baseline. The system is not yet customer-facing, and we conclude by discussing some of the work remaining.

\section{Related Work}

Recently, deep learning-based systems for question answering and dialogue have been the focus of both academic and industrial research. In dialogue systems, Vinyals et al \cite{vinyals15} and Serban et al \cite{serban15} demonstrated that encoder-decoder networks with LSTM units can generate dialogue based on IT help desk and movie script corpuses. For question answering problems, Sukhbaatar et al \cite{sukhbaatar15} are able to achieve competitive performance on the so-called bAbI tasks \cite{babi} with memory networks and limited supervision \cite{weston14}. Last year, Google launched Smart Reply, an email response recommendation system that recommends short replies for 10\% of Gmail volume in their Inbox mobile application \cite{smart-reply}. Google's new messaging app Allo also uses same technology to recommend responses for mobile chats.

In this paper, we applied a Siamese-like network \cite{siamese} with 2 encoders to build a response generation system for a subset of customer service chats related to item delivery problems. (In particular, we selected chat contacts where the customer indicated a ``Where's My Stuff?" issue.) In the context of information retrieval, Lowe et al \cite{ubuntu} also used a similar network to retrieve the next reply from a corpus of Ubuntu technical help IRC chats.

Our approach is most similar to that described in Smart Reply, but with certain differences. Firstly, while Smart Reply and our system both generate replies from a fixed set of templates, we do not need to perform beam search or generate the text directly. This simplifies the engineering effort required for deployment and speeds up response generation, since our responses can be longer than the ~10 token limit that the Inbox UI suggests. Secondly, while we both use clustering techniques and manual inspection to extract an initial set of templates, we perform the clustering in a fully automatic fashion, without the need for intent clusters initialized with human expertise. Clustering around pre-specified intents is important for an open domain corpus like emails, since there would be a huge number of topic clusters in the dataset, whereas in closed domains this is less important.

The paper is organized as follows: Section \ref{sec_dual_enc} provides details about the dual encoder network model. Section \ref{sec_templates} discusses how we create the pool of template answers that our dual encoder network model selects from. Section \ref{sec_eval} presents evaluations of our system and Section \ref{sec_conclusion} discusses future improvements.

\section{Models for Response Generation}
\subsection{Dataset}

We used one year of Amazon customer service chat transcripts on item delivery issues from 09/2015 - 09/2016 for creating training data. The raw text is split into agent and customer turns, tokenized, filtered for sensitive customer information (e.g. names, credit card numbers, etc.), and converted to lowercase.

The data needed for training the dual encoder network are pairs of customer questions and agent responses with binary labels of whether or not they are a match. 

To extract meaningful question-answer pairs, we select every customer turn in the conversation that ends with a question mark; the agent turn after it is considered the correct reply. These matching pairs constitute our positive samples.

To create non-matching pairs (i.e. negative samples), we use the same set of customer questions, but for each question, we randomly select an agent turn that follows some other customer question in the corpus. We created 3.3 million training samples with a positive to negative ratio of 1:2 in our training dataset. This dataset was extracted from a small fraction of the total contact volume we handled that year. Table \ref{data_examples} shows some positive and negative examples from our training data.

\begin{table*}[h]
	\caption{Sample training data pairs. To generate negative samples, we paired questions with randomly sampled responses.}
	\begin{center}
		\begin{tabular}{|c|c|c|}
			\hline
			Customer Inquiry & Agent Response & Label \\
			\hline
			and will i be sent an email ? & yes , NAME . & 1  \\
			can the ship speed be changed ? & yes , i 've already upgraded .& 1 \\
			ok so what i have to do now ?& it 's a good company to work for & 0 \\
			can I ask for a resend ? & both the orders will be delivered to you today . & 0 \\
			\hline
		\end{tabular}
	\end{center}
	\label{data_examples}
\end{table*}

\subsection{Dual Encoder Network}\label{sec_dual_enc}

\begin{figure}[h]
	\centering
	\includegraphics[width=8.75cm]{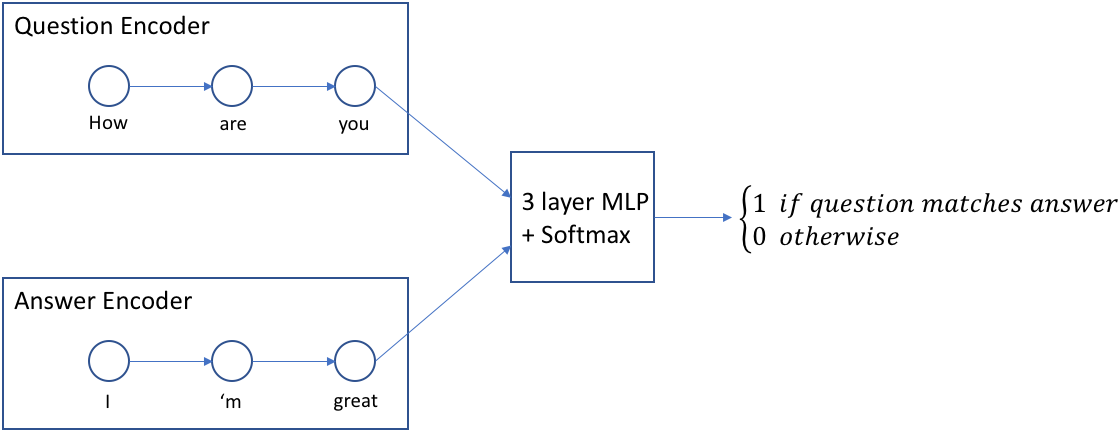}
	\caption{Schematic of the dual encoder network.}
	\label{network_diagram}
\end{figure}

Figure \ref{network_diagram} shows the schematic for the network.

Our dual encoder network takes a customer question (e.g. ``Will I receive a new tracking number?") and an answer (e.g. ``Yes we'll have it emailed to you.") as input. The question and answer are fed into two separate LSTM \cite{lstm} encoders. The encoders generate low dimensional embeddings for the question and the answer. The embeddings are then concatenated and passed to a multi-layer perceptron (MLP) which outputs the probability that the question and answer match.

LSTM networks have been widely used for encoding sentences into low dimensional embeddings for various NLP-related tasks. \cite{graves12} showed that LSTM's achieved state-of-the-art performance on various sequential classification tasks. Presently, LSTM-based classifiers are standard baselines for text classification tasks. \cite{sutskever14} applied LSTM's to create sentence embeddings for machine translation. \cite{kiros15} and \cite{wieting15} showed that LSTM-based embeddings can be used for transfer learning across diverse tasks, including semantic relatedness, paraphrase extraction, and information retrieval. In this work, we used LSTM's for encoding question and answer sentences, as shown in Figure \ref{network_diagram}. At time $t$, the word $w_t$ is mapped to a embedding $v_t$ and then fed into the LSTM one at a time, updating the hidden state $h_t$ of the LSTM. The hidden state of the LSTM at the last time step is used as the embedding of the entire sentence.

We trained the dual encoder model with Keras and Theano \cite{theano}. Among the hyperparameter combinations we tried, the optimal error on the development set was obtained with the hyperparameters listed in Table \ref{hyperparam}. We used Adam \cite{adam} to perform the stochastic optimization of the network parameters. The network has a total of ~5 million parameters.

\begin{table}[!htbp]
	\caption{Hyperparameters for the dual encoder network.}
	\label{tab:word_styles}
	\centering
	\begin{tabular}{|l|l|}
		\hline
		Hyperparameter      & Value         \\
		\hline
		Word embedding dim            & 512               \\
		LSTM output layer dim                   & 512      \\
		\# of MLP layers                    & 3         \\
		MLP hidden dim.                    & 3         \\
		MLP activation                    & ReLU         \\
		Learning rate                    & 0.0002         \\
		\# of epochs                    & 4         \\
		\hline
	\end{tabular}\label{hyperparam}
\end{table}

This model achieves $81\%$ accuracy on the development set, where the positive to negative ratio is also 1:2.

\subsection{Response Template Extraction and Prediction}\label{sec_templates}

The other key idea of the system is the pool of pre-constructed answer templates. An ideal pool would contain all of the common agent responses on item delivery issues; if the appropriate answers are not in the pool, then the system cannot recommend a reasonable answer. On the other hand, the pool size can't be too large due to the computational cost. While a pool of 10k randomly sampled agent answers will cover almost all common questions on item delivery issues at prediction time, the dual encoder network would have to score 10k question-answer pairs for each input customer question.

As the first step, we randomly sampled 400k agent answers from historical item delivery-related chats, and generated embeddings for them by using the trained answer encoder (Figure \ref{network_diagram}). Our analysis shows the embeddings are able to capture semantic similarity beyond simple vocabulary overlap (Table \ref{enc_examples}). The answer templates are selected with the help of K-means clustering. We applied mini-batch k-means with k-means++ initializations \cite{sculley2010, k-means} to cluster the 400k answer embeddings into 500 clusters. To form the template for each cluster, we take the text of the agent answer with embedding closest to the cluster center. Finally, we created a pool of 200 answer templates by human review.

At prediction time, the system will pair the customer question with every pre-constructed answer template, and use the trained dual encoder network to produce a measure of how well each answer matches the question. The system will then recommend the top-k answers to the agents ranked by this probability. Note that the answer embeddings for the full set of the answer templates are precomputed and stored for computational efficiency.
 
A more straightforward approach to performing dialogue response generation would essentially be a supervised text classification task. Based on the customer intent predicted by the model, the system can present the customer with some pre-determined response. A common example of this would be pre-written dialogue combined with state tracking, which is used in IVR systems in travel and restaurant reservation applications \cite{dstc}. However, even for something as simple as item delivery issues, the total number of possible types of customer questions can be close to 100-200, and extending such an approach to all of the domains in customer service (e.g. Kindle content, Amazon Instant Video issues, Prime subscription issues, etc.) would be impractical. In contrast, our approach has the benefit of not needing annotated data. The only data needed to train our model are customer questions and the agent answers after them, which already exist in our historical chat transcripts.

\section{Experiments}\label{sec_eval}

\subsection{Selected Examples}

The dual encoder produces 2 sets of embeddings: one for customer questions and another for agent answers. Table \ref{enc_examples} shows the 5 nearest neighbors for a few selected questions in this embedding space. We also present the 5 nearest neighbors for some selected answers. In contrast to the nearest neighbors found with tf-idf vectorization, LSTM embeddings seem to capture more semantic similarity, since tf-idf is essentially based on search term overlap. For example, LSTM embeddings find various kinds of responses to customer greetings even when the search terms do not overlap very much (e.g. the model finds both ``A pleasure to meet you too!" and ``Glad to hear that!"), while tf-idf only finds ones that share tokens in common. 

\begin{table*}[h]
	\caption{Nearest neighbors of questions and replies. The second column is based on embeddings from the question encoder and the reply encoder. The third column is based on the search results retrieved from the corpus based on tf-idf.}
	\begin{center}
		\tiny
		\begin{tabular}{|c|c|c|}
			\hline
			Question or Answer & Encoder nearest neighbors & tf-idf nearest neighbors \\
			\hline
			ok you start the refund ? & \makecell{ is the refund done ? \\
				is the full refund done already ? \\
				just to be clear , you have issued a refund for the original order ? \\
				did the refund already go through ? \\
				but you 've already done a refund ?} & \makecell{ should i start a new chat ?\\
				do you want me to start then ?\\
				i will try . how to i start ?\\
				how would you like to start ?\\
				and start the new return ?} \\
		    \hline
		    a pleasure to meet you too !&\makecell{ glad to hear that !\\
		    	it is our pleasure to assist our customers\\
		    	great ! !\\
		    	my pleasure : )\\
		    	fantastic !} & \makecell{ nice to meet you !\\
		    	nice to meet you : )\\
		    	nice to meet you .\\
		    	it 's a pleasure .\\
		    	it was a pleasure to meet and assist you :}\\
	    	\hline
		\end{tabular}\label{enc_examples}
	\end{center}
\end{table*}

\subsection{Answer Ranking}

We compare the dual encoder network with the tf-idf baseline on an answer ranking task. For this task, we paired 10k randomly sampled customer questions with the correct answer and 9 randomly sampled incorrect answers. In this task, the ``correct answer" is simply the agent response, which is not templatized. For each question, the 10 answers were ranked based on the probabilistic output of the dual encoder network. This ranking was compared to the one produced by the sum of the tf-idf term weights. We compare the mean reciprocal rank and precision@3 for both algorithms in Table \ref{prec_at_3}. On all metrics, the dual encoder network significantly outperforms the tf-idf baseline. We also present examples of customer questions and the matching answers in Table \ref{matching_qa_pairs}.

\begin{table*}[t]
	\caption{Customer questions and the answer templates recommended by our system}
	\begin{center}
		\tiny
		\begin{tabular}{|c|c|}
			\hline
			Question & Top 3 recommended answers \\
			\hline
			when will i receive my shoes ?&\makecell{ it will be delivered DATE \\
				you will get the items on DATE \\
				you 'll receive the package within 24 hours .
			} \\
			\hline
			how can i use the gift card balance ?&\makecell{ you can use it on your next purchase .\\
				you can use after 2 hours . because it will take only 1-2 hours to credit in your account .\\
				the refund will be reflected in your gift card balance in the next 1-3 hour}   \\
			\hline
			hi are you there ?&\makecell{yes I 'm here .\\
				yes , i 'm checking it .\\
				sorry for the delay in responding}   \\
			\hline
			can i cancel the order ?&\makecell{ i can cancel it for you .\\
				i 've cancelled it .\\
				which items you need to cancel ?}   \\
			\hline
			why it has n't been shipped yet ?&\makecell{ i am glad to check the status of your order .\\
				your order is already entered to the shipping process .\\
				it is out of stock.}   \\
			\hline
		\end{tabular}
	\end{center}
	\label{matching_qa_pairs}
\end{table*}

The tf-idf baseline does not perform well on this task because even when the vocabulary overlap contains signal for retrieving the answer given the question (e.g. both the question and answer contain ``gift card"), there are many cases where the answer to a question will not have any overlap at all (e.g. yes or no-type questions).

\begin{table}[!htbp]
	\caption{Results on answer ranking task. MRR refers to mean reciprocal rank.}
	\begin{center}
		\begin{tabular}{|c|c|c|}
			\hline
			  & MRR & Precision@3 \\
			\hline
			Dual encoder & 0.685 & 80.0\%\\
			tf-idf& 0.562 & 65.2\%\\
			\hline
		\end{tabular}
	\end{center}
	\label{prec_at_3}
\end{table}

\subsection{End-to-end Human Evaluation}

We recruited a rotating pool of agents to evaluate how well the system works end-to-end. We randomly selected 100 questions, and used our system and the tf-idf baseline to each recommend 3 answers (e.g. top 3 most probable answers from the 200 answer templates). The agent is asked to go through the question-answer pairs and assign a relevance score from 1 to 3 to each answer, with 3 being very relevant, 2 being somewhat relevant, and 1 being irrelevant. The evaluations are done on the same 100 questions for both algorithms.

\begin{figure}[h]
	\centering
	\includegraphics[width=8.5cm]{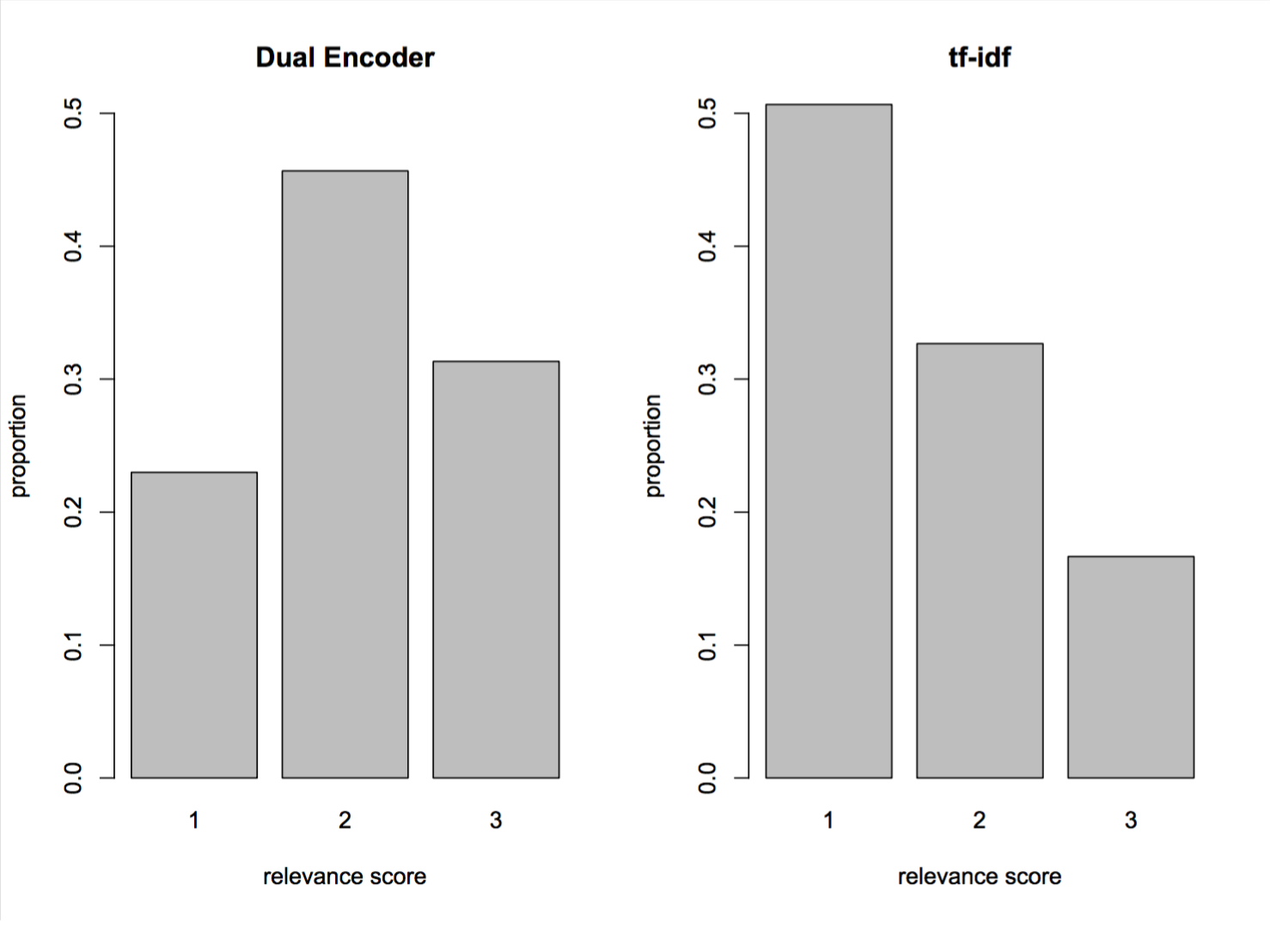}
	\caption{Human evaluation results based on relevance scores.}
	\label{relevance_scores}
\end{figure}

Note that given a question there maybe more than one appropriate answer. For example, answers like ``I'm sorry you can't" and ``Yes, you can cancel it from your order page" are both very relevant answers to the question ``Can I cancel the order since it's late?". Table \ref{matching_qa_pairs} shows sample model-based answers to questions, and Figure \ref{relevance_scores} shows the human evaluation relevance score distribution for both our system and tf-idf. In general, our system shows more high relevance recommendations compared with the tf-idf baseline, where $>$70\% of the model-selected templates are relevant to the question being asked.

Another metric we examined is within the top three answers, how often is there at least one ``very relevant" answer. Among the 100 randomly sampled questions, our system is able to recommend at least one very relevant answer (score = 3) among the top 3 for 48 questions, while the tf-idf baseline does so for only 31 questions. 

The average relevance score for the tf-idf model is 1.66 ($\pm$ 0.10, 95\% CI), whereas the average relevance score for the dual encoder baseline is 2.08 ($\pm$ 0.09, 95\% CI).

\section{Conclusion and Future Work}\label{sec_conclusion}

We have shown that a template-based approach to dialogue response generation works well in the customer service domain. We demonstrate that even in the absence of a fully automated dialogue system, it is nonetheless possible to select highly relevant answers to customer questions, which can translate into a reduction in the time spent per customer contact. Though we are currently testing this system for online chats, we believe that the template-based approach would extend naturally to a speech-driven system for telephone conversations.

There are a number of future directions we will pursue to make the system more complete.  We would like to determine the correct polarity for a given template based on the state of a customer's orders. For example, if a customer is inquiring about a shipment that has not yet arrived, we can reply either that the shipment is expected to be on-time or late from internal shipment data. We would also like to rerank the list of suggested templates using customer context, and expand the set of slots in our templates that can be filled automatically by internal systems.

\section{Acknowledgments}

The authors would like to thank Kevin Small for the helpful discussions and assistance in reviewing the drafts. We would also like to thank the customer service associates who helped us evaluate our system and provided domain-specific feedback on its design.

\bibliographystyle{IEEEtran}
\bibliography{mybib}

\end{document}